\documentclass{amia}
\usepackage{graphicx}
\usepackage{subcaption}
\usepackage[labelfont=bf]{caption}
\usepackage[superscript,nomove]{cite}
\usepackage{color}
\usepackage{adjustbox}
\usepackage{url}
\usepackage{makecell}
\usepackage{float}

\begin{document}

\title{Development of deep learning algorithms to categorize free-text notes pertaining to diabetes: convolution neural networks achieve higher accuracy than support vector machines}

\author{Boyi Yang, PhD$^{1, 2}$, Adam Wright, PhD $^{1, 2}$}

\institutes{
    $^1$Harvard Medical School;
    $^2$Brigham and Women’s Hospital\\
}

\maketitle

\section*{}
Keywords: Deep learning, Convolutional Neural Network, Recurrent Neural Network, Natural Language Processing, Electronic Health Record, Diabetes\\ \\

\noindent{\bf ABSTRACT}

\textit{\textbf{Background}
Health professionals can use natural language processing (NLP) technologies when reviewing electronic health records (EHR). Machine learning free-text classifiers can help them identify problems and make critical decisions. We previously introduced a support vector machine (SVM) classifier that can identify EHR progress notes pertaining to diabetes. Yet deep learning algorithms have also been shown to be effective in text classification. \\ \\
\textbf{Objective}
To develop and experiment with deep learning neural network algorithms that identify EHR progress notes pertaining to diabetes and validate the algorithms at two institutions.
\\ \\
\textbf{Materials and methods}
The data used are 2,000 EHR progress notes retrieved from patients with diabetes at the Brigham and Women's Hospital (1,000 for training and 1,000 for testing) and another 1,000 notes from the University of Texas Physicians (for validation). All notes were annotated manually as diabetic or non-diabetic. Several deep learning classifiers were developed, and their performances were evaluated with the area under the ROC curve (AUC).
\\ \\
\textbf{Results}
Among the deep learning models we experimented, the convolutional neural network (CNN) with a separable convolution layer accurately identified diabetes-related notes in the Brigham and Women's Hospital testing set with the highest AUC of 0.975. The AUC for the external University of Texas Faculty Physicians validation set is 0.875. 
\\ \\
\textbf{Discussion}
The CNN model was accurate and can be generalized to notes from another institution, whereas Recurrent Neural Networks (RNN) might not be suitable for binary classification of highly-specialized EHR notes.
\\ \\
\textbf{Conclusions}
Deep learning classifiers can be used to identify EHR progress notes pertaining to diabetes. In particular, the CNN-based classifier can achieve a higher AUC than an SVM-based classifier. \\ \\}
\\
\section*{INTRODUCTION}
With the advent of electronic health records, clinicians now have access to more data than ever before. Although this access has many positive benefits, clinicians face challenges reviewing large volumes of data, and need tools for summarizing and filtering information.\cite{feblowitz2011summarization} A particular challenge is summarizing and filtering free-text information, such as clinical notices.

We previously introduced an SVM approach to classify free-text notes pertaining to diabetes and achieved reasonable accuracy and generalization. Clinicians can use such classifier to selectively filter patients’ progress notes and identify those pertaining to diabetes.\cite{wright2013use} However, the existing algorithm is fairly simple (using bag of words and SVM), and we suspected that advances in deep learning for natural language processing (NLP), such as CNN and RNN, can be applied to the same free-text dataset and show better performance. 

Deep learning has previously been used in the diagnostic process and to assist physicians, especially in computer vision. There are abundant examples of deep learning applications for medical images. For instance, Gulshan et al. (2016) have completed a study on the detection of diabetic retinopathy with retinal fundus photographs,\cite{gulshan2016development} and Wang, J et al. (2016) on the discrimination of breast cancer with mammography.\cite{wang2016discrimination} However, the number of deep learning studies regarding medical texts is much less extensive compared to those with applications on medical images. One notable work is the study of CNN on medical text by Hughes et al. (2017), the model for which is included in our experiment with our dataset.\cite{hughes2017medical} Nonetheless, deep learning has made huge advances in general NLP; CNN and RNN have emerged as two widely used architectures and are often combined with tree-structured or sequence-based NLP models.\cite{zhou2015c} Yet, most text classification studies used publicly available benchmarks to measure the performance of their models but not directly on real clinical text data: Coneau et al. (2017) used Yelp Review and Yahoo! Answers.\cite{conneau2016very} Zhou et al. (2016), Wang et al. (2016), and Zhou et al. (2015) used the Stanford Sentiment Treebank.\cite{zhou2016text, wang2016discrimination, zhou2015c}

Our study is designed to bridge the gap between state-of-the-art deep learning models and real physicians’ EHR notes. We experimented with a variety of CNN, RNN, and hybrid models, and we introduced our unique models that were modified based on our experiments. Instead of using popular NLP benchmarks, we applied our models directly on clinical data: a large number of real EHR notes annotated by multiple physicians. We also validated our deep learning models at two sites and reported performance characters such as AUC. 
\\
\section*{METHODS}

\textbf{Creation and Annotation of Data Samples}

This study is a continuation of the free-text notes categorization study by Wright et al. (2013), and we employed the same data samples: a training sample and a test sample from Brigham and Women's Hospital (BWH), Boston, Massachusetts, USA, and a validation sample from the University of Texas Faculty Physicians (UTP), Houston, Texas, USA. The study was reviewed and approved by the Partners HealthCare Human Subjects Committee and The University of Texas Health Science Center at Houston Committee for the Protection of Human Subjects.\cite{wright2013use}

To develop the training and test datasets, we randomly chose from BWH 1,000 patients whose medical problem list showed an entry of diabetes mellitus, and then we retrieved 2,000 outpatient EHR notes from those 1,000 patients. The notes were manually annotated as either positive (the note addressed diabetes directly or mentioned a prior history of diabetes) or negative (the note had no mention of diabetes). The specifications of the annotation process were detailed by Wright et al. (2013) in their previous work with this dataset. After annotation, we divided the 2,000 notes randomly into two sets: 1,000 notes were assigned to the training sample and 1,000 notes to the test sample.\cite{wright2013use}

For the validation dataset, we retrieved 1,000 EHR outpatient notes from UTP, a hospital with patient and provider populations entirely distinct from those of BWH. Therefore the validation dataset can genuinely attest to the generalizability of our algorithm. These 1,000 notes were annotated as either diabetic or non-diabetic using the same scheme as for the training and test datasets.\cite{wright2013use}

\textbf{Tokenization and Encoding}

The 2,000 free-text notes (1,000 for training and 1,000 for testing) from BWH physicians were tokenized into 2,000 vectors of words. Then, a word-to-number index map was generated using the training set. The number of words in this index map is 20,218. The 2,000 word arrays from both the training and testing sets were then encoded using the index map, and each note was turned into an integer sequence. The max length for all notes is 4,932, while the 99 percentile length is only 1,532. Therefore, the integer sequences shorter than 1,532 were padded to 1,532 integers with zeros, and sequences longer than 1,532 were truncated. After encoding and padding, the training dataset and the test dataset became two 2D integer tensors of shape (1000, 1532) with their entries ranging from 0 to 20218. 

\textbf{Word Embedding}

We applied an embedding layer to the training dataset before applying any neural network. A good word-embedding space depends heavily on the task.\cite{chollet2017deep} Since physicians' free-text notes are highly specialized texts, we did not use pre-trained embedding but learned a new embedding space. An embedding layer was added so that the network learns a 12-dimensional embedding for all 20,218 words in the index map. The embedding dimension was set to 12 following a rule of thumb that states the embedding dimension should be the 4th root of the vocabulary size (20218), according to Google Tensorflow developer updates,\cite{tensorflow_team_2017} and the 2D integer tensors were turned into 3D floating point tensors of shape (1000, 1532, 12). Such 3D tensors can then be processed by the CNN or RNN layer. 

\textbf{Recurrent Neural Networks}

The first type of network we employed was the Recurrent Neural Network (RNN), as it has shown good results for text classification.\cite{lee2016sequential, liu2016recurrent} We started with a single LSTM layer as our baseline, and then we added dropout and additional LSTM layers in an attempt to increase the accuracy. We also explored the bidirectional LSTM (BLSTM), as it can offer higher performance than a regular RNN for natural-language processing tasks.\cite{zhou2016text, liu2016learning}

\begin{itemize}
	\item a. LSTM: Our baseline RNN model contained only one LSTM layer followed by the sigmoid activation function.
	\item b. LSTM with dropout: Recurrent Neural Networks such as LSTM generally have the problem of overfitting. Alternatively, dropout was applied to the input and recurrent connections of the memory units with LSTM precisely and separately. LSTM-specific dropout can have a more pronounced effect on the convergence of the network than the layer-wise dropout.\cite{chollet2017deep} In our study, all dropout rates were kept at 0.2 for subjective comparison. 
	\item c. 2-Stack LSTM with dropout: It has been shown that the performance of the RNN model can increase with the depth of LSTM layers.\cite{conneau2016very} Thus, we stacked two LSTM layers with same dropout input and recurrent dropout.
	\item d. 3-Stack LSTM with dropout: We stacked our model with another LSTM layer. Recurrent layer stacking is a classic way to build more powerful RNNs: for instance, the Google Translate algorithm is powered by a stack of seven LSTM layers.\cite{chollet2017deep} Thus, we increased the capacity of the network by adding yet another LSTM layer to understand whether this improved performance.
	\item e. Bidirectional LSTM with dropout: After AUC diminished as we stacked multiple LSTM layers (See Results section), we replaced the regular LSTM layer with one bidirectional LSTM (BLSTM) layer. RNNs are notably order-dependent. They process the encoded notes following the order in which the notes were written by the physicians, and shuffling or reversing the order can completely change the representations they extract from the sequence.\cite{chollet2017deep} By processing the text notes in both directions, the bidirectional RNN can catch patterns that may be overlooked by a unidirectional RNN.
\end{itemize}

\textbf{Convolutional Neural Networks}

We then experimented with CNN. Originally invented for computer vision, CNN models have also been shown as effective for NLP and have achieved excellent results in semantic parsing \cite{yih2014semantic}, search query retrieval,\cite{shen2014learning} sentence modeling,\cite{kalchbrenner2014convolutional} and other traditional NLP tasks.\cite{chollet2017deep} Here, we adopted the single layer CNN model by Kim et al. (2014) and the deep CNN model by Hughes et al. (2017).\cite{kim2014convolutional, hughes2017medical} Then, following each model, we present our modified version that improves both the accuracy and running speed on the training data. 

\begin{itemize}
	\item f. CNN with dropout: We started building our CNN with the simple CNN model proposed by Kim et al. This CNN model has one convolutional layer followed by a max pooling layer and a dropout layer. We chose this model as our baseline model since it has been shown that CNN with one layer of convolution performs remarkably well for text classification.\cite{kim2014convolutional}
	\item g. Separable CNN with dropout: We then modified the simple CNN model by replacing the regular convolutional layer with a separable convolutional layer since adding a separable filter in the convolution model allows substantial speedups with a possible increase in accuracy.\cite{jaderberg2014speeding}
	\item h. Deep CNN: A study by Huges et al. (2017) has demonstrated that a deep CNN model can have excellent accuracy in medical text classification. We employed Huges et al.’s best performing multi-layer CNN on our embedded data. This model consisted of a configuration of two sets of two convolutional layers, and each pair was followed by a max pooling layer. After the second max pooling function, a dropout function was applied to help prevent overfitting. Then, a fully connected layer was appended followed by a second dropout function.\cite{hughes2017medical} The model ended with another single neuron dense layer with a sigmoid activation function.
	\item i. Deep CNN simplified: In this last CNN model, we modified Hughes et al.’s deep CNN to reduce computation cost. Each set of convolutional layers was cut down to just one convolution layer, while all other layers were kept the same. 
	
\end{itemize}

The architectures of our modified model g and model i are presented in Figure 1.
\begin{figure}[H]
	\centering
	\begin{subfigure}[b]{.5\textwidth}
		\centering
		\includegraphics[width=.8\linewidth]{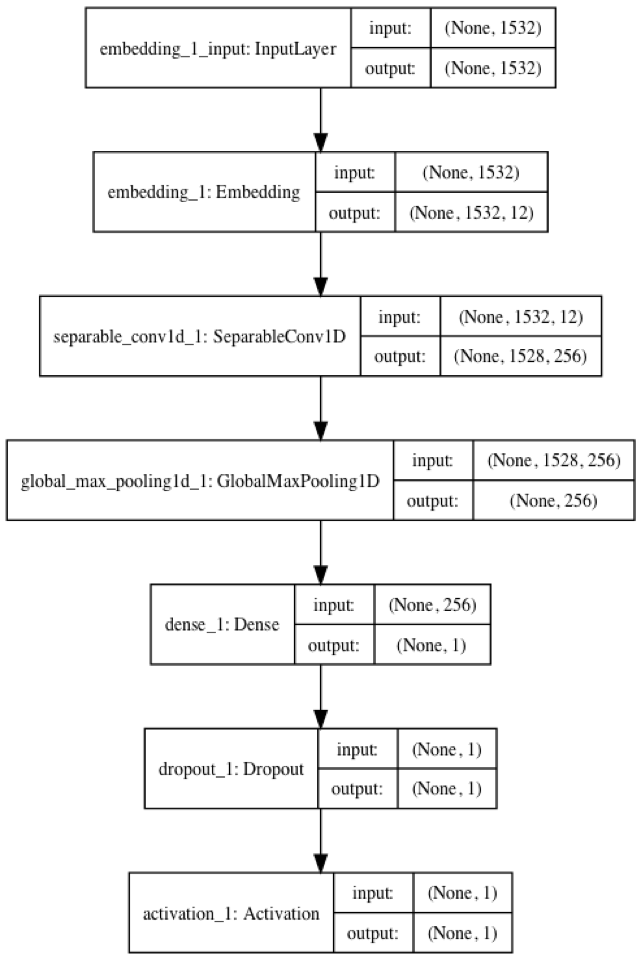}
		%\caption{1a}
		%\label{fig:sfig1}
	\end{subfigure}%
	\begin{subfigure}[b]{.45\textwidth}
		\centering
		\includegraphics[width=.8\linewidth]{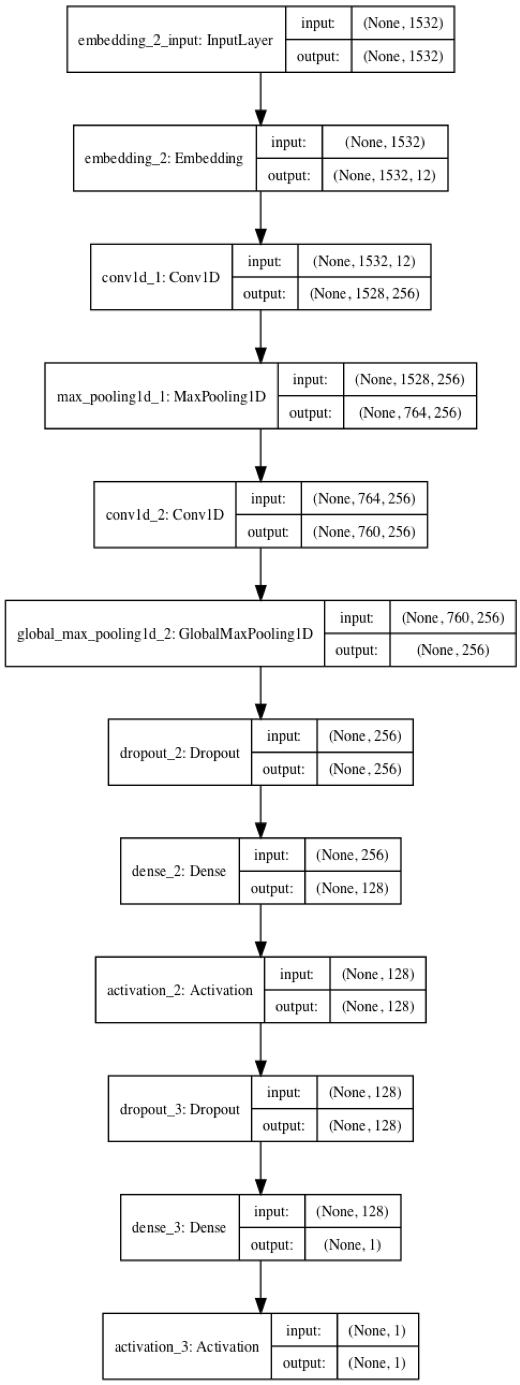}
		%\caption{1b}
		%\label{fig:sfig2}
	\end{subfigure}
	\caption{Architectures of CNN model g (left) and CNN model i (right)}
	\label{fig1}
\end{figure}

\textbf{Hybrid Neural Network}

We then created a hybrid approach by adding a LSTM after the CNN. This strategy combines the speed and lightness of CNNs with the order-sensitivity of RNNs and could be beneficial since long sequences that cannot realistically be processed with RNNs.\cite{chollet2017deep} The combination of CNN and LSTM can create synergy and outperform both CNNs and LSTMs.\cite{zhou2015c} We then modified our baseline hybrid model by replacing LSTM with BLSTM and by swapping the order of CNN and RNN.

\begin{itemize}
	\item j. Hybrid CNN-LSTM: we added a one-dimensional CNN and max-pooling layers after the embedding layer, which then fed the consolidated features to the LSTM. We took this model as our baseline hybrid model.
	\item k. Hybrid CNN-BLSTM: Bidirectional LSTM is the best performing RNN for our data. We modified our baseline hybrid model by replacing the regular LSTM layer with a bidirectional LSTM layer. 
	\item l. Hybrid BLSTM-CNN: Last, we swapped the positions of the LSTM layer and CNN. This model extracted sequential features in both forward and backward directions and output features to a convolutional layer. 
\end{itemize}

The architectures of our modified model k and model l are presented in Figure 2.

\begin{figure}[H]
	\centering
	\begin{subfigure}[b]{.5\textwidth}
		\centering
		\includegraphics[width=.8\linewidth]{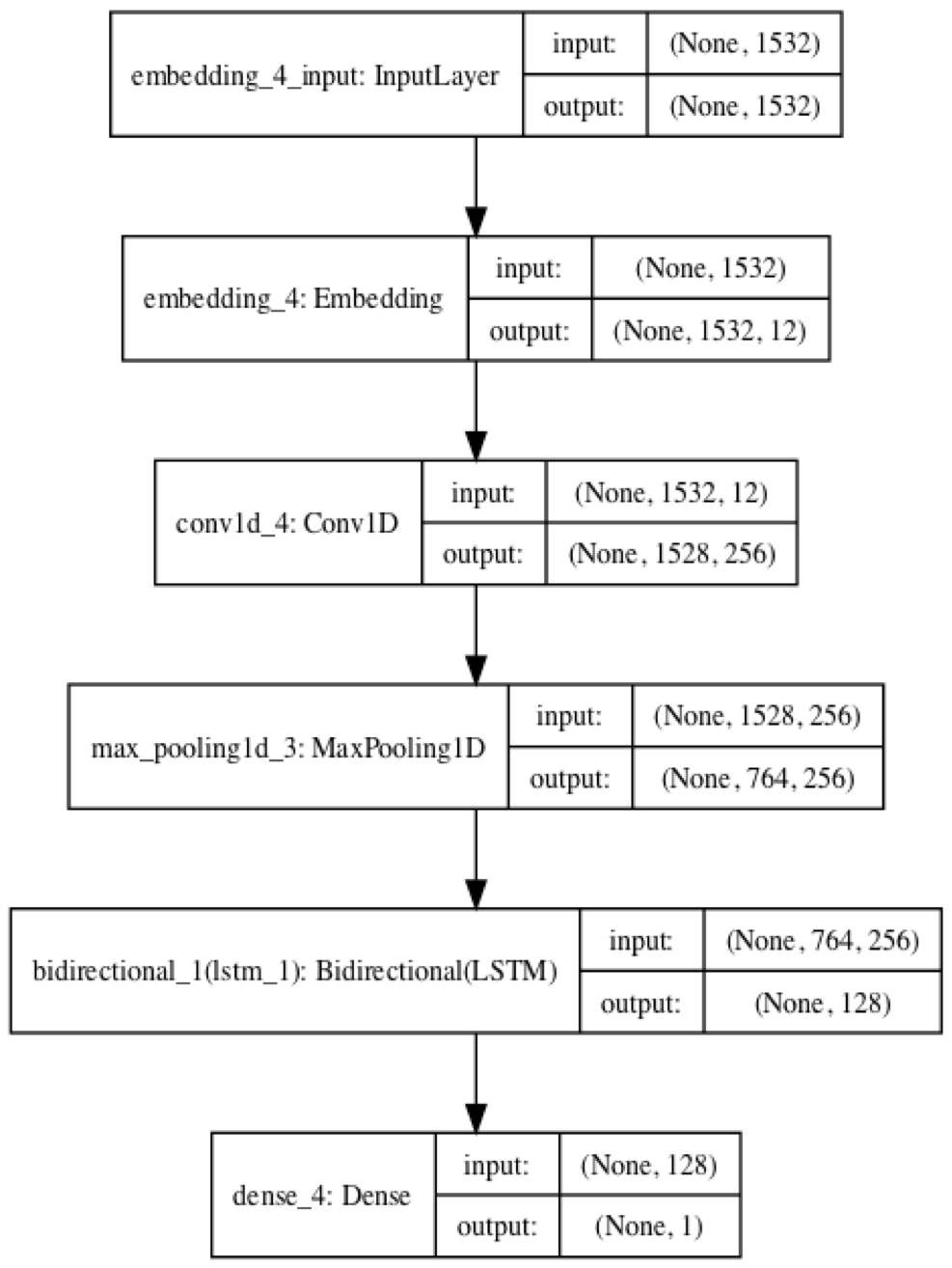}
		%\caption{1a}
		%\label{fig:sfig1}
	\end{subfigure}%
	\begin{subfigure}[b]{.45\textwidth}
		\centering
		\includegraphics[width=.8\linewidth]{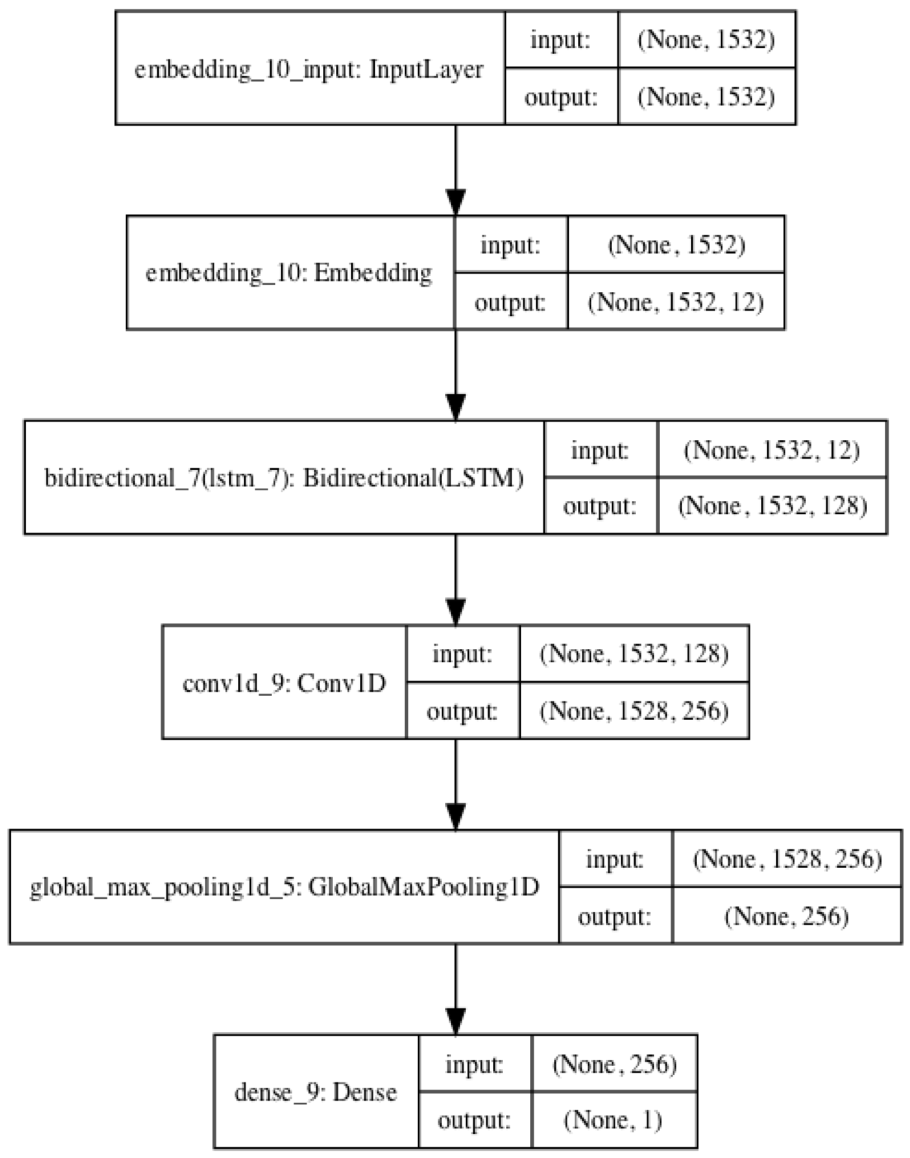}
		%\caption{1b}
		%\label{fig:sfig2}
	\end{subfigure}
	\caption{Architectures of Hybrid model k (left) and Hybrid model l (right)}
	\label{fig2}
\end{figure}

Our deep learning models were built using Keras, an open source neural network library designed to enable fast experimentation with deep neural networks and capable of running on top of TensorFlow.\cite{chollet2017deep, abadi2015tensorflow} Keras gets its source of randomness from a NumPy random number generator.\cite{oliphant2006guide} In addition, TensorFlow also has its own random number generator. To have the same initial weights for subjective comparison of different neural networks, both generators were set by the same seeds (seed 1 for NumPy and seed 2 for TenserFlow) for all the models.
\\
\section*{RESULTS}
We monitored the training loss, and we stopped when the training loss did not reduce more than 0.01 for two consecutive epochs. Then, we computed the AUC of the epoch with the lowest training loss (two epochs before the stop). The results are presented in Table 1. 

\begin{table}[H]
	\centering
	\begin{adjustbox}{width=1\textwidth}
		\begin{tabular}{|l|l|l|l|l|}
			\hline
			\thead{Network Type}   & \thead{Model}  & \thead{Test AUC \\ on sample from BWH} & \thead{Validation AUC \\ on sample from UTP}  & \thead{Number of Epochs \\ to reach minimum loss}  \\ 
			\hline
			SVM  &  SVM with bag of word & 0.956  & 0.947 & NA   \\ 
			\hline
			RNN & a. LSTM (experiment)  & 0.877& 0.798 &   19 \\ 
			\cline{2-5}
			& b. LSTM with dropout (experiment)  & 0.850& 0.770 &   22 \\ 
			\cline{2-5}
			& c. Stacked LSTM 2-layer (experiment) & 0.833& 0.751 &   14 \\ 
			\cline{2-5}
			& d. Stacked LSTM 3-layer (experiment)  & 0.818& 0.742 &   17 \\ 
			\cline{2-5}
			& e. Bidirectional LSTM (experiment)  & 0.883& 0.795 &   21 \\ 
			\hline
			CNN & f. CNN with dropout (Kim et al. (2014))  & 0.972& 0.873 &   15 \\ 
			\cline{2-5}
			& g. Separable CNN with dropout (our modification) & 0.975& 0.875 &   16 \\ 
			\cline{2-5}
			& h. Deep CNN (Hughes et al. (2017)) & 0.940 & 0.870 &   8 \\ 
			\cline{2-5}
			& i. Deep CNN Simplied (our modification) & 0.964 & 0.872 &   8 \\ 
			\hline
			Hybrid  &  j. CNN-LSTM (Zhou et al. (2015)) & 0.864  & 0.788 & 19   \\ 
			\cline{2-5}
			&  k. CNN-BLSTM (our modification) & 0.872  & 0.790 & 20   \\
			\cline{2-5} 
			&l. BLSTM-CNN (our modification) & 0.926 &0.841 & 14\\
			\hline
		\end{tabular}
	\end{adjustbox}
	\caption{Results of all deep learning models we experimented}
	\label{table1}
\end{table}

The best RNN model achieves test AUC of 0.88 and did not surpass the SVM approach. RNN might not be suitable to categorize free-text notes, at least not for binary classification of highly specialized text such as EHR notes from physicians. RNN models achieve a lower AUC than CNN models. Contrary to our expectations, adding more LSTM layers further decreases the AUC. The best performing RNN is the bidirectional LSTM when considering both the forward and backward order of the word sequences, yet it only achieves an AUC of 0.88. 

CNN models have the best AUC. Not only do CNN models outperform any pure RNN model, they also achieved a higher validation AUC than the SVM approach. CNN is also efficient on our dataset – one simple CNN layer with dropout already reached an AUC of 0.972. Deep CNN with 4 convolutions did not improve AUC. However, separable convolution gives us an edge both in speed and accuracy.
The separable CNN model with dropout (Model g) achieved both the highest test AUC (0.975) and the highest validation AUC (0.875), and its accuracy and loss values up to the stopping epoch are presented in Figure 3. The test and validation ROC curves of Model g are shown in Figure 4.

\begin{figure}[H]
	\centering
	\begin{subfigure}[b]{.45\textwidth}
		\centering
		\includegraphics[width=.8\linewidth]{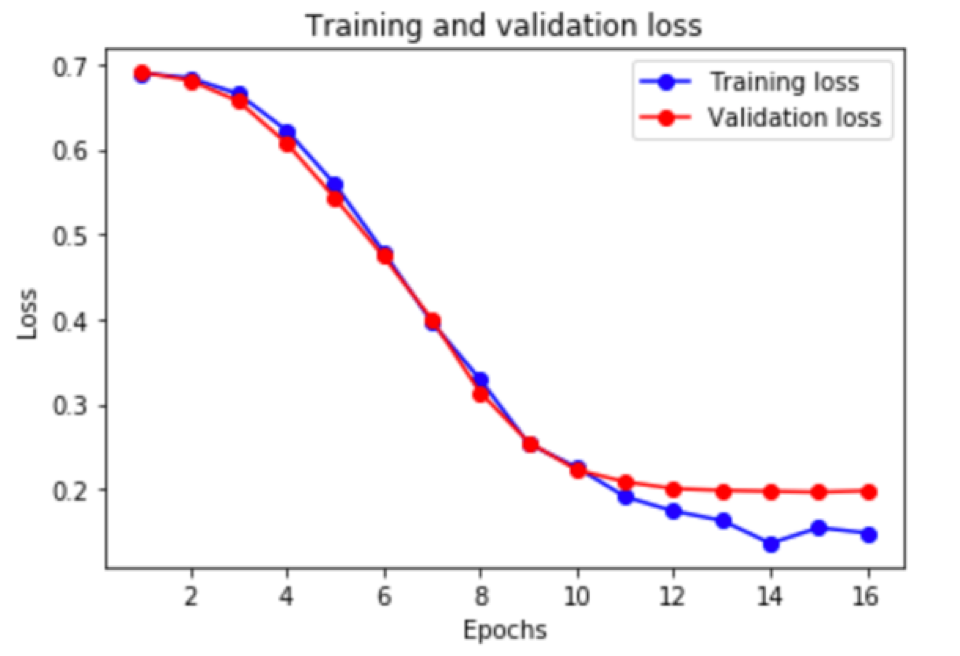}
		%\caption{1a}
		%\label{fig:sfig1}
	\end{subfigure}%
	\begin{subfigure}[b]{.45\textwidth}
		\centering
		\includegraphics[width=.8\linewidth]{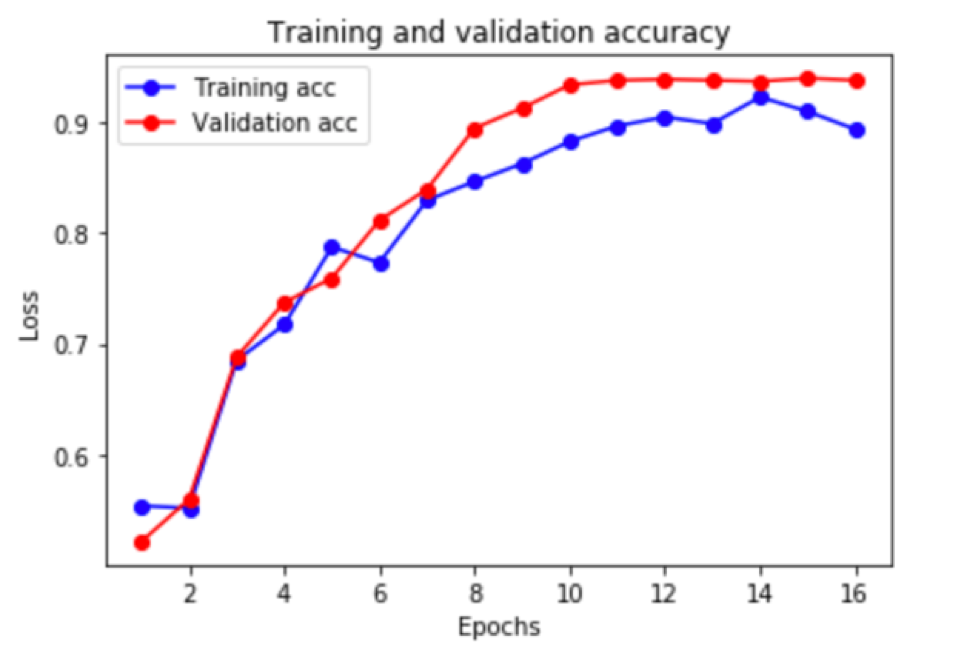}
		%\caption{1b}
		%\label{fig:sfig2}
	\end{subfigure}
	\caption{Accuracy (left) and Loss (right) of separable CNN (Model g)}
	\label{fig3}
\end{figure}

\begin{figure}[H]
	\centering
	\includegraphics[scale=0.6]{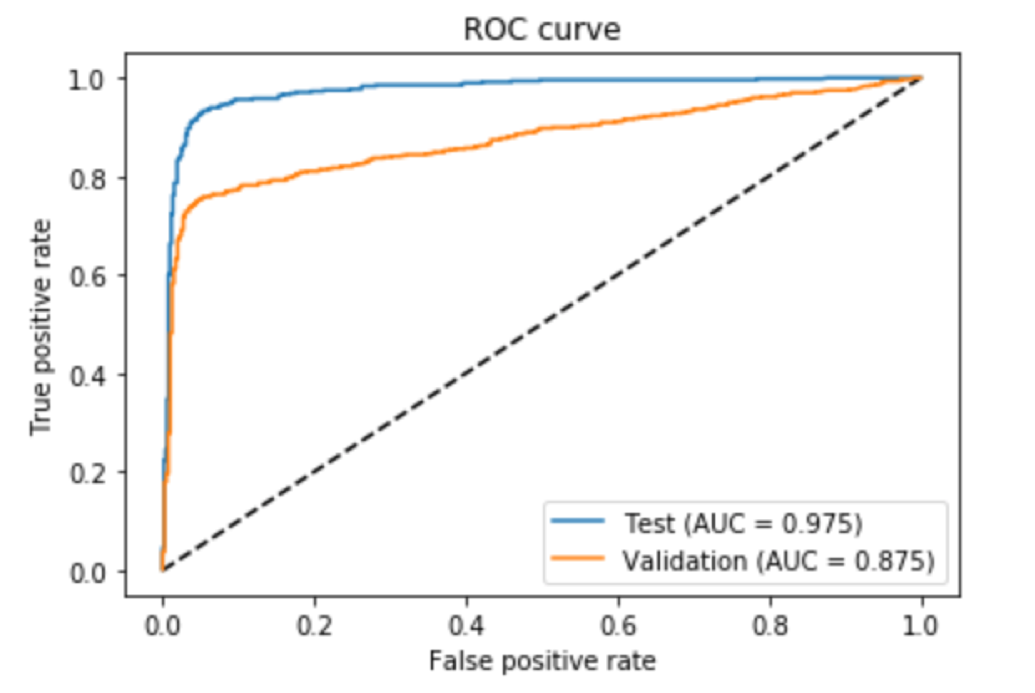}
	\caption{ROC curves of separable CNN (Model g)}
	\label{fig4}
\end{figure}

Adding an LSTM layer after the CNN does not show improvement. On the other hand, positioning a bidirectional LSTM in front of the CNN gave great performance but still did not surpass our CNN models. For our free-text word-embedded dataset, the combination of CNN and RNN does not show any synergy. A pure CNN is the champion of all models we tested.
\\
\section*{DISCUSSION}
Our deep learning models generally worked well with our EHR notes. All models achieved test AUC higher above 0.80. In particular, CNN models achieve higher test AUC than the SVM approach, suggesting that deep learning models can recognize sentence patterns beyond the mere count of certain high-frequency words. CNN models’ higher test AUC within BWH can be attributed to that physicians in one institution might have a particular pattern that CNN can capture. When an EHR free-text categorization tool is employed on the dataset from the same institution as the training dataset, a deep learning approach—especially CNN—should be considered over SVM.

The validation AUC of deep learning models is consistently lower than the validation AUC of SVM. Given that our validation dataset is from another hospital, the lower validation AUC implies that deep learning models might not generalize across multiple institutions as well as SVM. The low validation AUC at UTP reflects a tradeoff between accuracy and generalization: the CNN models captured patterns and note writing styles that are specific to BWH, resulting in their higher test AUC within BWH, but the unique patterns learned from BWH hurt the validation AUC at UTP. Therefore, if the users need to classify notes from multiple institutions but have only one large training set from one institution, then it is safe to just use SVM for the higher generalizability.

Should we still use deep learning classifiers to categorize EHR notes? The answer is yes. First, almost all CNN models (Model f, g, and i) outperformed SVM on test AUC, confirming that CNN has a small yet consistent advantage over SVM. Second, the CNN models with higher test AUC than SVM are fast. Especially, the separable CNN (Model g) took only 161 seconds to reach the stopping point (on a MacBook Computer, 2.3GHz 2-core Intel i5 Processor, 8GB RAM). Even though SVM is still more efficient than neural networks, with today’s computer power, the speed difference between separable CNN and SVM is practically negligible. Last, the deep learning classifier has the potential to improve in accuracy if we have more data for training. SVMs are effective for relatively small data sets with fewer outliers,\cite{chi2008classification, foody2006use} while deep learning shines with large datasets. As users acquire more data through the usage of the classifier, deep learning can reach higher than the 0.975 AUC that we obtained with only 1,000 training notes. Users should consider deep learning classifiers such as CNN if they desire maximal accuracy within one institution.

One limitation on our study is that our neural networks were applied only to word-embedded data: we encoded our notes only at the word level. Some studies have experimented with character embedding\cite{dos2014deep, zhang2015character} and sentence embedding.\cite{dos2014deep, liu2016recurrent} The performance of our deep learning models could change if we were to encode our notes at a different level. In addition, the code can produce slightly different results due to the machine precision, even though the weight for each model were set by the same seeds both at the NumPy and at the TensorFlow level. We acknowledge that the safest way to compare each model is to repeat each experiment at least 100 times and use summary statistics. Last, our study was conducted for only one disease. Whether a deep learning text categorization tool can be generalized to other medical conditions is yet to be investigated. In fact, we plan to expand deep learning EHR free-text note categorization to four more diseases in a subsequent study. 
\\
\section*{CONCLUSION}

Deep learning classifiers can be used to identify EHR progress notes pertaining to diabetes. In our diabetes dataset, CNN models have higher AUC than RNN or hybrid models. Simple CNN models performed the best: replacing regular convolution with separable convolution improves AUC and speed, yet increasing the number of convolution layers does not. In comparison, SVM generalizes better across different institutions, yet CNN models outperform SVM when employed within one institution. Future studies should focus on whether CNN models can be applied to other diseases.
\\
%\newpage
%\section*{Another Major Heading and References}
%This sentence has two reference citations\cite{ref1,ref2}.
%
%More text of an additional paragraph, with a figure reference (Figure ~\ref{fig1}) and a figure inside a Word text box below.  Figures need to be placed as close to the corresponding text as possible and not extend beyond one page.\\
%\begin{figure}[h!]
%\centering
%\includegraphics[scale=1]{pics/figure1.png}
%\caption{Total allergy alerts, overridden alerts, or drug order cancelled.}
%\label{fig1}
%\end{figure}

\makeatletter
\renewcommand{\@biblabel}[1]{\hfill #1.}
\makeatother

\bibliographystyle{unsrt}

\bibliography{References_Jamia.bib}

%\begin{thebibliography}{1}
%\setlength\itemsep{-0.1em}

%\bibitem{ref1}
%Pryor TA, Gardner RM, Clayton RD, Warner HR. The HELP system. J Med Sys. 1983;7:87-101.
%\bibitem{ref2}
%Gardner RM, Golubjatnikov OK, Laub RM, Jacobson JT, Evans RS. Computer-critiqued blood ordering using the HELP system. Comput Biomed Res 1990;23:514-28.

%\end{thebibliography}

\end{document}